%% file: root.tex
\documentclass[letterpaper, 10 pt, conference]{ieeeconf}  

\IEEEoverridecommandlockouts                              

\usepackage{comment}
\usepackage{amsmath}
\usepackage{amsfonts}
\usepackage{graphicx}
\usepackage{array}
\usepackage{multirow}
\usepackage{caption}
\usepackage{hyperref}
\usepackage{gensymb}

\title{\LARGE \bf
Continuous Wrist Control on the Hannes Prosthesis: \\a Vision-based Shared Autonomy Framework}

\author{Federico Vasile, Elisa Maiettini, Giulia Pasquale, Nicolò Boccardo and Lorenzo Natale%
\thanks{Federico Vasile, Elisa Maiettini, Giulia Pasquale and Lorenzo Natale are with the Istituto Italiano di Tecnologia, Humanoid Sensing and Perception, 16163 Genoa, 
Italy (email: \texttt{name}.\texttt{surname}@iit.it).}%
\thanks{Federico Vasile is also with the Dipartimento di Informatica, Bioingegneria, Robotica e Ingegneria dei Sistemi (DIBRIS), University of Genova, 16146 Genoa, Italy.}%
\thanks{Nicolò Boccardo is with the Istituto Italiano di Tecnologia, Rehab Technologies, 16163 Genoa, 
Italy, and also with the Open University Affiliated Research Centre at Istituto Italiano di Tecnologia (ARC@IIT), 16163 Genoa, Italy (email: nicolo.boccardo@iit.it) .}%
\thanks{The Open University Affiliated Research Centre at Istituto Italiano di Tecnologia (ARC@IIT) is part of the Open University, Milton Keynes MK7 6AA, United Kingdom.}%
\thanks{This work was partially supported by the Istituto Nazionale Assicurazione Infortuni sul Lavoro, under the project iHannes (PR19-PAS-P1), the European Union’s Horizon-JU-SNS-2022 Research and Innovation Programme under the project TrialsNet (Grant Agreement No. 101095871) and the project RAISE (Robotics and AI for Socio-economic Empowerment) implemented under the National Recovery and Resilience Plan, Mission 4 funded by the European Union – NextGenerationEU.}%
}

\usepackage{eso-pic} 
\newcommand{\firstpagecopyright}{%
  \AddToShipoutPictureFG*{%
    \AtPageUpperLeft{%
      \hspace*{\dimexpr1in+\oddsidemargin\relax}%
      \raisebox{-3.5\baselineskip}[0pt][0pt]{%
        \begin{minipage}{\textwidth}
          \centering\footnotesize
          \textit{\textcopyright~2025 IEEE. Personal use of this material is permitted.
          Permission from IEEE must be obtained for all other uses, in any current or future media,
          including reprinting/republishing this material for advertising or promotional purposes,
          creating new collective works, for resale or redistribution to servers or lists, or reuse of
          any copyrighted component of this work in other works.}\\
          Preprint version (Feb.\ 2025). This work has been accepted for publication in ICRA 2025.
        \end{minipage}%
      }%
    }%
  }%
}

\begin{document}
\firstpagecopyright   
\maketitle
\thispagestyle{empty}
\pagestyle{empty}

\begin{abstract}
Most control techniques for prosthetic grasping focus on dexterous fingers control, but overlook the wrist motion. This forces the user to perform compensatory movements with the elbow, shoulder and hip to adapt the wrist for grasping. 
We propose a computer vision-based system that leverages the collaboration between the user and an automatic system in a shared autonomy framework, to perform continuous control of the wrist degrees of freedom in a prosthetic arm, promoting a more natural approach-to-grasp motion. Our pipeline allows to seamlessly control the prosthetic wrist to follow the target object and finally orient it for grasping according to the user intent. We assess the effectiveness of each system component through quantitative analysis and finally deploy our method on the Hannes prosthetic arm. Code and videos: \url{https://hsp-iit.github.io/hannes-wrist-control}.

\end{abstract}

\section{INTRODUCTION}
\label{sec:introduction}
\input{Sections/introduction}

\section{RELATED WORK}
\label{sec:related_work}
\input{Sections/related_work}

\section{CONTROL ARCHITECTURE}
\label{sec:method}
\input{Sections/method}

\section{DATASETS}
\label{sec:datasets}

\input{Sections/datasets}

\section{EXPERIMENTS}
\label{sec:experiments}
\input{Sections/experiments}

\section{APPLICATION ON THE HANNES PROSTHESIS}
\label{sec:application}
\input{Sections/application}

\section{CONCLUSIONS}
\label{sec:conclusions}
\input{Sections/conclusions}


\bibliographystyle{IEEEtran}
\bibliography{bibliography}

\end{document}

%% file: Sections/introduction.tex
    Latest advancements in prosthetic technologies have made significant steps toward restoring motor functions of amputees. However, achieving dexterous and intuitive control of prosthetic hands is yet a challenging task, requiring to meet the user intent with the complicated joints motion needed for object grasping. Most commercial prostheses are based on \textit{electromyography} (EMG) or \textit{mechanomyography} (MMG), relating these input signals to the velocity of the prosthesis motors~\cite{chen2023}. When more than one degree-of-freedom (DoFs) is available, the Sequential Switching and Control (SSC) paradigm is used. In this case, only one joint at a time is driven and the user gives an explicit input signal to switch between the different DoFs, resulting in a cumbersome control~\cite{amsuess2014}. Therefore, relieving the user from complex control input modalities is of high interest in prosthetics. 
Exploiting the know-how from grasp synthesis techniques in robotics~\cite{newbury2023deep} might be a possibility. However, they generally predict a set of grasp poses, then naively select one (e.g., highest score) to be executed by the robot. Instead, in prosthetics, the prediction must conform to the user's intent, i.e., how the user is approaching the object.
In view of this, the \textit{shared-autonomy} (or \textit{shared-control}) principle has been introduced in the literature~\cite{gardner2020}, relying on the collaboration between the user and a semi-autonomous system, generally exploiting additional sources of input such as images or inertial measurements. However, previously presented semi-autonomous systems~\cite{vasile2022,starke2022} do not consider a continuous control for the prosthesis during the approach-to-grasp action. Instead, we believe that in order to foster a more natural grasping approach, the automatic system should continuously drive the joints in compliance with the user motion.

\begin{figure}
    \vspace{+0.3cm}
    \centering
    \includegraphics[width=1.0\linewidth]{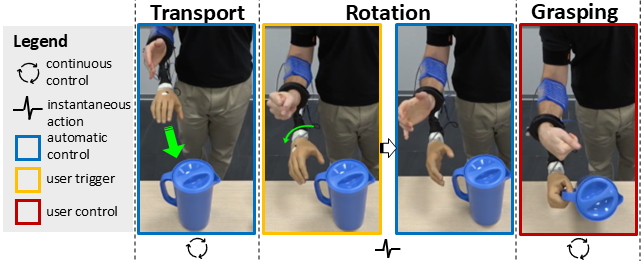}
    \caption{The phases of the prosthetic grasping pipeline.}
    \label{fig:phases}
    \vspace{-0.6cm}
\end{figure}

In this work, we introduce a novel \textit{eye-in-hand} vision-based shared autonomy system designed to continuously control the wrist DoFs of a prosthetic arm.
We present a prosthetic grasping pipeline based on three phases (see Fig.~\ref{fig:phases}): (i) first, while the user approaches the object, an automatic system continuously control the prosthetic wrist to follow the object in order to achieve a natural motion (\textit{transport} phase); (ii) then, as soon as the user triggers a signal, the system predicts the target object part and prepares the wrist accordingly for grasping (\textit{rotation} phase); (iii) finally, the control is left to the user who will use the EMG signals to drive the fingers opening-closing (\textit{grasping} phase). Moreover, we propose an object parts segmentation network, called DINOv2Det, that exploits the powerful feature descriptors of DINOv2~\cite{oquab2023}, together with the well-established Mask R-CNN~\cite{he2017} structure for instance segmentation. We devised a tool to generate a synthetic dataset for objects parts segmentation and a semi-automatic pipeline to annotate existing prosthetic grasping video sets. We used these for training and testing our vision system. We tested each component of the proposed \textit{eye-in-hand} shared autonomy control pipeline with datasets and simulation and, finally, we deployed it on the Hannes~\cite{laffranchi2020hannes} arm to verify its effectiveness on a real prosthesis.


%% file: Sections/related_work.tex
\noindent{\textbf{Prosthetic control}.} While the SSC paradigm is a well-established control method, it prevents the direct control of multiple DoFs simultaneously.
A viable alternative and active field of research is the EMG \textit{pattern recognition}. This aims to recognize repeatable and distinct features from the EMG input signals and associate them to the hand movements. However, bringing these techniques into the real world is still a challenge due to robustness issues (e.g., electrode shift, muscle fatigue, etc.)~\cite{chen2023}. Another possibility is exploiting additional input sources. For instance, in~\cite{castro2022semi}, the user aims at the target object with the prosthesis and four laser scanner lines are used to estimate the object shape and size. Then, the wrist orientation and grasp size are automatically computed. In~\cite{shi2023hand}, instead, the information of the hand-object pose during the approach-to-grasp action is leveraged to predict the final wrist pose and pre-shape type. However, the model needs the hand-object pose as input, requiring a preliminary pose estimation. 
Other approaches exploit the visual input from an \textit{eye-in-hand} camera (i.e., a camera embedded into the prosthesis palm) as additional information, showing promising results~\cite{vasile2022, starke2022, robotics12060152, shi2022target}. 
For instance, in~\cite{shi2022target}, an object detector is used to identify the target object in the clutter and predict the grasp type. Similarly, in~\cite{starke2022}, the embedded camera is used to take a picture of the target object, classify it, retrieve the associated grasp from a database and suggest it to the user. The database is also exploited to select the best hand and wrist pre-grasp trajectories for the approach. However, taking a picture before grasping may hinder a smooth and natural approach to grasp. Thus, in most of the systems for transradial amputees, either the hand is not controlled during the movement or the control is non-adaptive—meaning it relies on a one-shot decision before grasping rather than continuous control during the approach. 
Instead, we propose a control architecture for transradial amputees that integrates visual servoing~\cite{chaumette2006} for continuous wrist control during the approach, enabling more natural movement. Additionally, we segment the target object into parts and utilize this finer information to predict the final wrist configuration for grasping.

\noindent{\textbf{Object parts segmentation}.} Segmenting the object into parts for prosthetic grasping might resemble the affordance detection problem as both computer vision tasks aim at identifying the different parts of an object and the human actions (or grasp types) associated to them. However, they are substantially different and this limits the adoption of affordance detection systems for prosthetic grasping. Specifically, the affordance detection task is typically framed with a coarse labeling~\cite{nguyen2017,myers2015} using, e.g., only the class \textit{grasp}, while, in prosthetics, specific grasp types are required. Furthermore, available image datasets are mostly taken from egocentric or external cameras while the \textit{eye-in-hand} configuration is preferable in prosthetics~\cite{vasile2022}. Finally, the affordance recognition is generally framed as an object detection followed by a semantic segmentation of the affordances within the bounding box~\cite{do2018affordancenet}. 
Instead, in prosthetic grasping, it is fundamental to identify the different parts of an object sharing the same affordance label as distinct instances.
Therefore, in this work, we frame the task as an instance segmentation since we need to consider all object parts as separate instances and select one of them to obtain the final wrist configuration. Finally, to overcome the aforementioned limitations of affordance detection datasets, we devise a semi-automatic procedure to obtain ground truth masks for a video dataset from prior work~\cite{vasile2022} and we propose a synthetic dataset generation tool. We use this data for training and testing the models.



%% file: Sections/method.tex
We propose an \textit{eye-in-hand} vision-based pipeline to control the wrist of a prosthetic hand using a camera embedded into the palm. The framework is presented in Sec.~\ref{sec:shared_autonomy_prosthetic_grasping}, and its components are detailed in Sec.~\ref{sec:visual_servoing_control} and Sec.~\ref{sec:object_parts_segmentation}.

\subsection{Shared-autonomy prosthetic grasping}
\label{sec:shared_autonomy_prosthetic_grasping}

A reach-to-grasp task can be split into three different phases: \textit{transport}, \textit{rotation} and \textit{grasping} (or \textit{manipulation})~\cite{gentilucci1991}. During the \textit{transport} stage, the user moves the hand toward the target object. Right before the hand closure, the wrist should be aligned according to the object part to grasp (\textit{rotation} step). Finally, in the \textit{grasping} phase, the fingers close around the object. 

The proposed work aims to closely follow the above phases in order to enable a natural grasping approach. We delegate the \textit{transport} and \textit{rotation} phases to an automatic vision system while leaving only the final action (i.e., \textit{grasping} phase) to the user, thus reducing the cognitive load. 
Specifically, throughout the \textit{transport} phase, given an input image, an object parts segmentation network locates the object, then a visual servoing control scheme drives the wrist to keep the object in the image center. As a result, the wrist is continuously adjusted in accordance with the user's movements around the object. However, it might end up in a suboptimal configuration for grasping. Therefore, when the user triggers the \textit{rotation} phase through an explicit EMG signal, the visual servoing stops running and the wrist is instantaneously configured for grasping using a prediction based on object parts. Moreover, note that since the visual servoing kept the object in the field of vision, an optimal view is ensured, resulting in better grasp prediction. Finally, the control is left to the user who uses the EMG signals to drive the fingers opening and closing to grasp the object (i.e., \textit{grasping} phase). We refer to Fig.~\ref{fig:phases} for a detailed illustration.
The presented framework, based on a \textit{shared-autonomy} paradigm, relieves the user from complex mode switching and control of multiple DoFs. 

\subsection{Visual servoing control}
\label{sec:visual_servoing_control}
In this section, we first revise the traditional visual servoing control. 
Then, we discuss how the prosthetic setting differs from the standard robotic setup. We leverage such argument in Sec.~\ref{sec:visual_servoing_simulation} to design an ad hoc control method for the Hannes~\cite{laffranchi2020hannes} prosthetic arm.

\noindent{\textbf{Visual servoing background}.} A classical visual servoing control system minimizes the error function $\mathbf{e}(t) = \mathbf{s}(t) - \mathbf{s}^*$,
being $\mathbf{s}(t)$ and $\mathbf{s}^*$ the current and target visual features, respectively. We design a velocity controller in the joint space~\cite{chaumette2007} for an \textit{eye-in-hand} system. 
Hence, the control law is:

\vspace{-0.2cm}
\begin{equation}  \label{eqn:control}
\dot{\mathbf{q}} = - \lambda \: (\mathbf{L}_s \: ^c\mathbf{V}_e \: ^e\mathbf{J}_e(\mathbf{q}))^+ \: (\mathbf{s}(t) - \mathbf{s}^*)
\end{equation}

where $\mathbf{q} \in \mathbb{R}^n$ encodes the position for $n$ robot joints, $^e\mathbf{J}_e(\mathbf{q}) \in \mathbb{R}^{6\times n}$ is the robot \textit{Jacobian} expressed in the end-effector frame, $^c\mathbf{V}_e \in \mathbb{R}^{6\times6}$ is the \textit{spatial motion transform matrix}~\cite{chaumette2006} to transform velocities from the end effector to the camera frame, $\mathbf{L}_s \in \mathbb{R}^{k\times6}$ is the \textit{interaction matrix}~\cite{chaumette2006}, $\mathbf{L}^+$ is the Moore-Penrose pseudo-inverse of a matrix $\mathbf{L}$ and $\lambda$ is the visual servoing gain. The features $\mathbf{s}(t)$, $\mathbf{s}^*$ and the \textit{interaction matrix} $\mathbf{L}_s$ depend on the chosen visual servo scheme.
We adopt the Image Based Visual Servoing (IBVS) due to its robustness under imprecise measurements~\cite{chaumette2006, corke2023}. 
Finally, given an input image, the IBVS will iteratively drive the robot joints $\mathbf{q}$ to align the current feature to the target. We refer to this scheme as \textit{standard} IBVS (s-IBVS).

\begin{figure}
    \centering
    \vspace{+0.30cm}
    \includegraphics[width=1.0\linewidth]{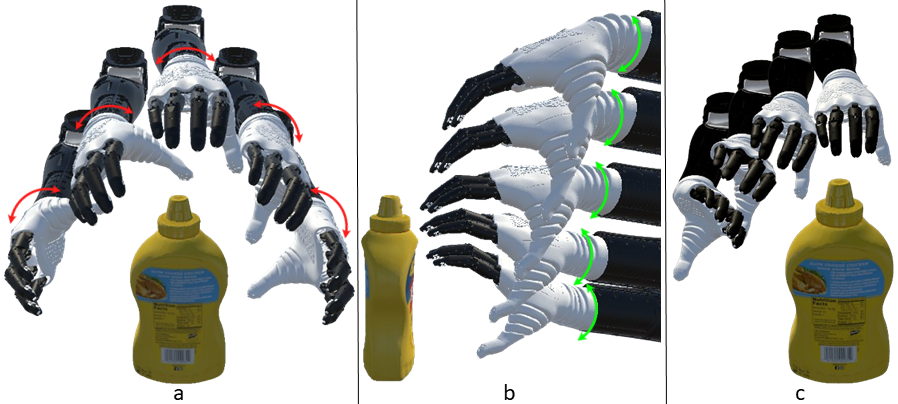}
    \caption{The natural wrist motion (a-b) and a non-natural motion (c) as the user drives the arm around the object.}
    \vspace{-0.60cm}
    \label{fig:ps_fe_angles}
\end{figure}
\newcolumntype{?}{!{\vrule width 1.5pt}}

\noindent{\textbf{Visual servoing for prosthesis control}.} While in a classical scenario the robot base is fixed and the control system drives the joints to bring the end-effector to the target, 
here the user moves the prosthetic hand while the robot joints (i.e., the prosthesis wrist) should perform compensatory motions to keep the object in the field of view. In such condition, given the strict coupling between the user movements and the control scheme, it is crucial to generate joint motions that are compliant with the user intentions (see Fig.~\ref{fig:ps_fe_angles}a-b). 
In this regard, as we will discuss in Sec.~\ref{sec:visual_servoing_simulation}, the s-IBVS might generate non-natural trajectories (see Fig.~\ref{fig:ps_fe_angles}c). This is mitigated by applying a similar idea to Partitioned IBVS~\cite{corke2023}: being $j$ the robot joint causing these trajectories, we remove it from s-IBVS and apply a separate control law for it. The objective is to obtain a natural motion while ensuring convergence. Thus, for this joint, a simple proportional control can be applied. The details are specified in Sec.~\ref{sec:visual_servoing_simulation} as they depend on the desired motion. Instead, the other joints are still controlled using the s-IBVS, thus $\mathbf{q} \in \mathbb{R}^{n-1}$ and $^e\mathbf{J}_e(\mathbf{q}) \in \mathbb{R}^{6\times {(n-1)}}$ in Eq.~\eqref{eqn:control}. We call this method \textit{proportional} and \textit{partitioned} IBVS (pp-IBVS).

\subsection{Object parts segmentation}
\label{sec:object_parts_segmentation}
We use the information of the object location in the image in two parts of the pipeline presented in Sec.~\ref{sec:application}. Firstly, we use the object mask centroid as input for the visual servoing during the \textit{transport} phase. In addition, the segmented object parts are used to predict the final wrist configuration during the \textit{rotation} phase.
Each object part is considered as a different instance and labeled with a grasp type (or \textit{no grasp} for non-graspable parts). Thereby, we propose to use an instance segmentation network. We remark that, differently from related work~\cite{starke2022}, the object parts are segmented based on the grasp types. We do not inject any object-specific information into our model as we strive for an object-agnostic pipeline by design. Even though the generalization to novel objects is not the focus of this work, we believe that an object-agnostic model would serve as a basis for such generalization, facilitating future research. 

\noindent{\textbf{DINOv2Det}.} For the proposed vision system, we leverage the recently introduced DINOv2~\cite{oquab2023} foundation model as a general-purpose backbone. Since DINOv2 is a Vision Transformer (ViT)~\cite{dosovitskiy2020}, it is not directly applicable for the instance segmentation task. Nevertheless,~\cite{li2022} proposed ViTDet, a network based on Mask R-CNN~\cite{he2017} structure but with a ViT as backbone. In addition, as opposed to the well-established Feature Pyramid solution, ViTDet only uses the last feature map of ViT to produce multi-scale feature maps. Then, the pipeline follows the standard Mask R-CNN: the multi-scale feature maps are used as input for the Region Proposal Network (RPN) and the Region of Interest (RoI) heads. The architecture obtains remarkable results when ViT is pre-trained in a self-supervised fashion. 
These findings support the strategy of using pre-trained ViTs as general purpose backbones, with minimal adaptation for the downstream tasks. Therefore, we push this idea forward by replacing the original ViT backbone of ViTDet with DINOv2. 
We refer to this novel architecture as DINOv2Det.
Notice that despite DINOv2 features have shown great capabilities in clustering semantic parts (e.g., differentiate between the legs and the body of an animal)~\cite{oquab2023}, our object parts do not necessarily have such distinction. 
Indeed, in some cases, no exact boundaries in terms of object shape or texture can be drawn (e.g., see the \texttt{010\_banana} in Fig.~\ref{fig:obj_parts}). We discuss it in details in Sec.~\ref{sec:exp_object_parts_segmentation}. 


%% file: Sections/datasets.tex

In the considered grasping task, the prosthetic hand might approach the object from any direction. Adopting an \textit{eye-in-hand} configuration, this means that the object parts segmentation model should work well from all the object viewpoints. 
However, collecting and labelling a dataset with such characteristic is a tedious procedure. Therefore, motivated by recent works on bridging the sim-to-real gap~\cite{vasile2022}, we aim to train the models on a synthetic dataset and evaluate them on a real dataset. 
Given the similarity with the task and setting, in this work, we adopt the real dataset from~\cite{vasile2022} but with significant modifications. Firstly, while in~\cite{vasile2022} the grasp type labels are defined in terms of the fingers configuration for grasping (e.g., \textit{pinch} grasp), in this work, we consider the wrist configurations (i.e., \textit{top grasp} and \textit{side grasp}). Therefore, we convert every object part label considered in~\cite{vasile2022} from the fingers to the wrist configuration (e.g., the \textit{power grasps} of the \texttt{006\_mustard\_bottle} are converted to \textit{side grasps}). Secondly, in~\cite{vasile2022} the label is assigned to an entire video, while for this study we consider a per-frame instance segmentation task. Thus, masks labels are required. In the following sections, we describe how we adapt the prosthetic dataset, namely the \textit{iHannes} dataset, collected in~\cite{vasile2022}, to the task at hand and the generation process of the synthetic dataset used for training the vision models.

\begin{figure}
    \centering
    \vspace{+0.30cm}
    \includegraphics[width=1.0\linewidth]{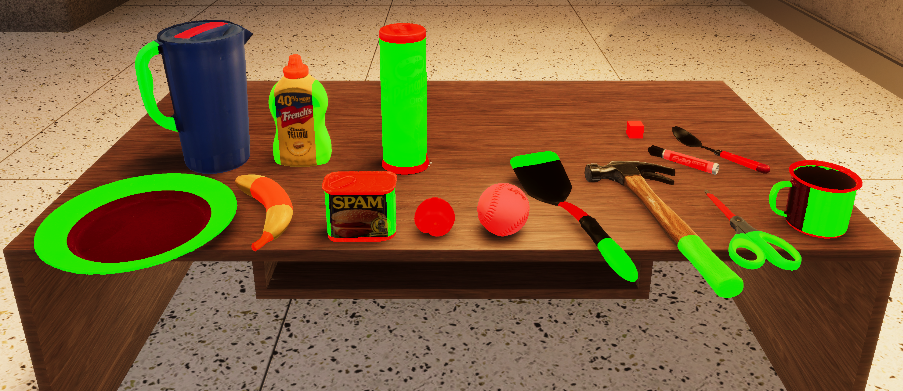}
    \caption{The 15 YCB objects with labeled object parts. The red and green masks encode the \textit{top} and \textit{side grasps}, respectively. 
    The non-highlighted object parts are labeled as \textit{no grasp}.  
    }
    \vspace{-0.60cm}
    \label{fig:obj_parts}
\end{figure}

\subsection{Real dataset annotation}
\label{sec:real_dataset_annotation}
The \textit{iHannes} dataset consists of RGB-D videos of approaching-to-grasp actions for 15 YCB objects~\cite{calli2015}. Since no ground-truth masks are available, in this work, we devise an efficient way to obtain them. Despite the emerging availability of semi-automatic mask annotation tools, the masks of the considered object parts may not have clear boundaries (e.g., a mask boundary is not necessarily aligned with the edges of the object's texture), resulting in error-prone manual labeling (i.e., incoherent parts boundaries between images). Therefore, we conceived a two step approach: (i) we manually partition the 3D mesh of all the 15 considered objects into graspable parts and assign the correspondent grasp type to each of them (see Fig.~\ref{fig:obj_parts}); then, (ii) for every image, the pose of the object ($o$) with respect to the camera ($c$), i.e. $\mathbf{T}_{c,o} \in SE(3)$, is used to project the object mesh onto the image plane in order to obtain the exact masks of the object parts. However, $\mathbf{T}_{c,o}$ is unknown and needs to be computed for every image.

In the \textit{iHannes} RGB-D videos, the target object position is fixed and the camera is moving toward it. Consequently, while the camera-to-object pose needs to be directly estimated for the first frame ($\mathbf{T}_{c^1,o}$), for any subsequent frame $k$, the relative camera displacement from frame $1$ to $k$ ($\mathbf{T}_{c^1,c^k}$) is sufficient to obtain $\mathbf{T}_{c^k,o}$. For instance, let $\mathbf{T}_{c^1,c^2}$ denote the camera pose from the first to the second frame (i.e., camera displacement) and $\mathbf{T}_{c^2,o}$ the camera-to-object pose for the second frame. $\mathbf{T}_{c^2,o}$ can be computed as $\mathbf{T}_{c^2,o} = \mathbf{T}_{c^1,c^2}^{-1} \mathbf{T}_{c^1,o}$. In general, being $\mathbf{T}_{c^{k-1},c^k}$ the camera displacement from frame $k-1$ to $k$, we first derive the camera displacement from frame $1$ to frame $k$ as $\mathbf{T}_{c^1,c^k} = \prod_{i=1}^{k-1}\mathbf{T}_{c^i, c^{i+1}}$ , then $\mathbf{T}_{c^k,o} = \mathbf{T}_{c^1,c^k}^{-1} \mathbf{T}_{c^1,o}$ is obtained. Hence, $\mathbf{T}_{c^1,o}$ and $\mathbf{T}_{c^i,c^{i+1}}$ are needed for every video.

In practice, we run a state-of-the-art object pose estimator~\cite{wang2021} on the first frame of the video to obtain a coarse estimate of $\mathbf{T}_{c^1,o}$ and we manually refine it to be used as ground truth. Then, the camera displacement for subsequent frames ($[\mathbf{T}_{c^i,c^{i+1}}]_{i=1}^{n-1}$) is estimated using~\cite{park2017colored}. We applied this two-step procedure rather than directly estimating $\mathbf{T}_{c^i,o}$ for every frame $i$ using~\cite{wang2021} since its predictions can be noisy (e.g., if the object is scarcely visible). Instead, estimation of the camera displacement~\cite{park2017colored} allows to use features from the surrounding environment. We used this procedure on the \textit{Same person} subset~\cite{vasile2022} of the \textit{iHannes} dataset, resulting in 311 videos labeled with the camera-to-object pose for every frame ($\mathbf{T}_{c^i,o}$). However, some frames were discarded because the blur caused a too noisy estimate of $\mathbf{T}_{c^i,c^{i+1}}$. Overall, 14692 frames were labeled with the $\mathbf{T}_{c^i,o}$ pose. Finally, note that in some cases it was not possible to initialize $\mathbf{T}_{c^1,o}$ using~\cite{wang2021}, since the pre-trained weights do not include a few objects used in the iHannes dataset. For those cases, the initialization of $\mathbf{T}_{c^1,o}$ was done by manually aligning the object mesh with the object in the point cloud.

\subsection{Synthetic dataset generation}
\label{sec:synthetic_dataset_generation}
While the \textit{iHannes} set is composed of real grasping videos and is used for testing the proposed prosthetic vision system, a training set comprised of images and mask labels is also required. Morever, it should meet the main requirement of the \textit{eye-in-hand} configuration, i.e., the variability of object viewpoints. Thus, we design a dataset generation tool.

We import the partitioned meshes of the 15 YCB objects, introduced in the previous section, in the \href{https://unity.com/}{Unity engine}. The data generation pipeline works as follows: (i) one object at a time is considered in the indoor tabletop scene; (ii) we uniformly sample 400 points on the surface of a upper hemisphere centered on the 3D centroid of the object; (iii) the simulated camera is placed on each of these points, look at the object and capture an image. We use the Unity High Definition Render Pipeline (HDRP) to obtain photo-realistic images and the Perception package to collect the ground-truth masks of the object parts. In addition, we apply domain randomization to various extent: (i) we randomize the background, light condition, table and position of the object; (ii) we use hemispheres of radius in the range from 0.2 to 1 meter by applying stratified sampling, i.e., the range is divided into 6 bins and then uniformly sample values within each bin; (iii) we apply random rotations (between 0 and 90 degrees) about the camera optical axis and make the camera look at a random point on the object. In summary, 400 points are sampled for each radius bin, resulting in 2400 images per object. This dataset is used for training the vision model. Moreover, one further dataset with slight modifications is generated for model validation. Specifically, for this, 100 points are sampled on the surface of the same hemisphere. In this case, the background and light condition are not randomized but we use a completely different scene, representing an outdoor environment with sunlight.

%% file: Sections/experiments.tex
\subsection{The Hannes prosthetic device}
\label{sec:the_hannes_prosthetic_device}
We test the proposed methods on the Hannes arm~\cite{laffranchi2020hannes}, considering the setup for a right arm transradial amputation~\cite{boccardowrist}. It has three DoFs: wrist flexion-extension (WFE), wrist pronation-supination (WPS) and fingers opening-closing (FOC). The WFE and WPS are revolute joints that are orthogonal and intersect at a common point (see Fig.~\ref{fig:trajectories}c). The FOC is a single DoF being the fingers actuated all together using one motor. Moreover, a tiny RGB camera is embedded into the prosthesis palm. It points downward with an angle $\theta = 16\degree $ with respect to the WPS rotational axis.

\subsection{Visual servoing simulation}
\label{sec:visual_servoing_simulation}

\noindent\textbf{Setup}. The virtual Hannes~\cite{di2021hannes} is imported into Unity to evaluate the visual servoing schemes in simulation. The aim is to control the WFE and WPS to bring the object at the image center. The IBVS represents the features as image plane coordinates, thus, we use the ground truth object mask from simulation and select the mask centroid as the current feature $\mathbf{s}(t)$, while the image center is the target $\mathbf{s}^*$. 
Hereafter, we firstly present an example highlighting the limitation of the s-IBVS and, then, how the pp-IBVS overcomes it. Finally, we provide a quantitative comparison.

\begin{figure}
    \vspace{+0.3cm}
    \centering
    \includegraphics[width=1.0\linewidth]{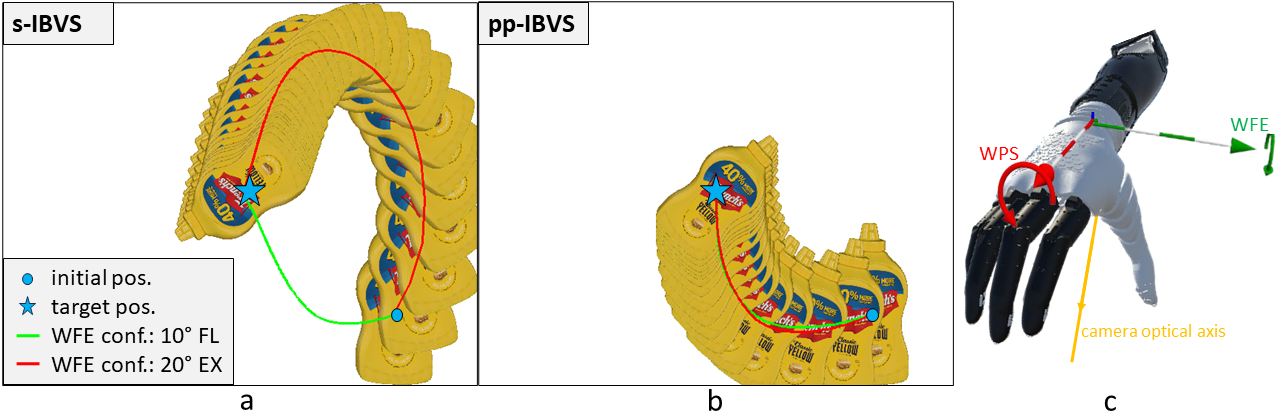}
    \caption{The trajectories generated by the visual servo schemes for two different WFE initial configurations (a-b). The Hannes arm (c).
    }
    \label{fig:trajectories}
    \vspace{-0.6cm}
\end{figure}

\noindent\textbf{Standard IBVS (s-IBVS)}. This controls both the WFE and WPS using Eq.~\eqref{eqn:control}. Our objective is to analyze the travelled trajectories in two scenarios having the same hand-object pose (hence same initial input image) but different initial WFE configurations (i.e., $10\degree$ flexion and $20\degree$ extesion). We run the control scheme while keeping the arm fixed in space and analyze the wrist motion. As shown in Fig.~\ref{fig:trajectories}a, the s-IBVS demands two completely distinct trajectories. The green-line trajectory is considered natural since the wrist rotates inward, i.e., the normal to the palm point toward the object (as in Fig.~\ref{fig:ps_fe_angles}a). Instead, the red-line trajectory is non-natural as the wrist rotates outward (similarly to Fig.~\ref{fig:ps_fe_angles}c). This is due to the different WFE initial configuration. Indeed, as the WFE rotates, the angle $\theta$ between the WPS rotational axis and the camera optical axis varies. In addition, when the WPS rotates, any point on the image plane translates both vertically and horizontally following a circular path governed by $\theta$. Therefore, there exist initial values of $\theta$ such that, in the first optimization step, a WPS outward rotation has the lowest error toward the target, eventually leading to a non-natural trajectory.

\noindent \textbf{Proportional and Partitioned IBVS (pp-IBVS)}. The s-IBVS controls both the WFE and WPS using Eq.~\eqref{eqn:control}, however, the WPS joint causes non-natural trajectories. Consequently, we control only the WFE using Eq.~\eqref{eqn:control} and apply a proportional control law for the WPS: $\dot{q}_{wps} = sign \: \lambda_{wps} \: | \mathbf{e}_{wps} |$, being $\lambda_{wps}$ the proportional gain. Then, at each timestep, we set $sign$ to either +1 or -1 depending on whether the object mask centroid is to the left or right of the image center and $\mathbf{e}_{wps} = x_c - x_o$ (i.e., the horizontal offset between the object mask centroid and the image center). This way, the direction of the WPS rotation only depends on the hand-object relative positioning. As a result, a WPS inward rotation is always executed (see Fig.~\ref{fig:trajectories}b), resulting in a natural wrist motion (as in Fig.~\ref{fig:ps_fe_angles}a).

\noindent\textbf{s-IBVS vs. pp-IBVS}. 
We aim to quantitatively assess the convergence time and naturalness of the final configuration through experiments in simulation. We sample 20 points on the surface of a upper hemisphere centered on the \texttt{006\_mustard\_bottle}. We place the hand on each point and randomly orient the wrist (i.e., the WFE and WPS) such that the object is in view. We run the control schemes and analyze both the convergence time (i.e, iterations required to converge) and the naturalness of final configuration (i.e., successful if the normal to the palm points toward the object). As shown in Tab.~\ref{table:vservo_quantitative}, the convergence time for pp-IBVS is slightly higher, this is expected since two different control laws are used for WPS and WFE, though the naturalness of WPS is ensured by design thanks to the ad hoc control law. Instead, since s-IBVS controls both WPS and WFE using Eq.~\ref{eqn:control}, the shortest path toward the target is ensured. However, this may result in non-natural wrist configurations.

\begin{table}
    \vspace{+0.2cm}
    \centering
    \renewcommand{\arraystretch}{1.2} 
    \caption{Quantitative Results for s-IBVS and pp-IBVS}
    \begin{tabular}{ccc} 
        \cline{1-3}
        Method & Convergence time (iter.) & Natural configuration (succ.) \\ \hline
        s-IBVS           & $\mathbf{213.5 \pm 124.9}$  & $13/20$ \\ 
        pp-IBVS          & $361.7 \pm 70.5$  & $\mathbf{20/20}$    \\ \hline
       
    \end{tabular}
    \label{table:vservo_quantitative}
    \vspace{-0.4cm}
\end{table}
    
\begin{table*}[ht]
    \centering
    \vspace{+0.3cm}
    \renewcommand{\arraystretch}{1.2} 
    \setlength{\tabcolsep}{5pt} 
    \captionsetup{justification=centering}
    \caption{Bounding Box and Segmentation Results on the Synthetic and Real Sets} 
    \begin{tabular}{p{3.35cm}?ccc|ccc?ccc|ccc?c} 
        \cline{1-14}
        \multirow{3}{*}{\textbf{Method}} & \multicolumn{6}{c?}{\textbf{Bounding box}} & \multicolumn{6}{c?}{\textbf{Mask}} & \multirow{3}{*}{\textbf{Inf. time (ms)}} \\
        \cline{2-13}
         & \multicolumn{3}{c|}{\textbf{Hemisphere val. set}} & \multicolumn{3}{c?}{\textbf{iHannes test set}} & \multicolumn{3}{c|}{\textbf{Hemisphere val. set}} & \multicolumn{3}{c?}{\textbf{iHannes test set}} \\ 
        \cline{2-13}
        & AP & AP$_{50}$ & AP$_{75}$ & AP & AP$_{50}$ & AP$_{75}$ & AP & AP$_{50}$ & AP$_{75}$ & AP & AP$_{50}$ & AP$_{75}$ \\ \hline

        Mask R-CNN~\cite{he2017} & \textbf{76.2} & 94.9 & 85.1 & 37.7 & 68.8 & 38.0 & \textbf{67.0} & 90.4 & \textbf{75.8} & 23.1 & 50.6 & 18.0 & \textbf{41.0} \\       
        ViTDet~\cite{li2022} & 73.3 & \textbf{96.0} & 82.9 & 43.7 & 78.7 & 43.6 & 64.1 & 91.3 & 72.6 & 29.1 & 62.0 & 24.0 & 128.2 \\ \hline        

        \hline
        
        DINOv2Det (Ours) & 76.1 & 95.6 & \textbf{85.2} & \textbf{55.7} & \textbf{91.4} & \textbf{60.7} & 66.9 & \textbf{91.7} & 75.2 & \textbf{36.9} & \textbf{73.8} & \textbf{32.9} & 54.6 \\  \hline     
        
    \end{tabular}
    \label{table:res_origtext}
\end{table*}

\begin{table*}[ht]
    \centering
    \vspace{+0.1cm}
    \renewcommand{\arraystretch}{1.2} 
    \setlength{\tabcolsep}{5pt} 
    \captionsetup{justification=centering}
    \caption{Texture Generalization Results When Training on Random Textures and Testing on Unseen Textures}
    \begin{tabular}{p{3.35cm}?ccc|ccc?ccc|ccc?c} 
        \cline{1-14}
        \multirow{3}{*}{\textbf{Method}} & \multicolumn{6}{c?}{\textbf{Bounding box}} & \multicolumn{6}{c?}{\textbf{Mask}} & \multirow{3}{*}{\textbf{Inf. time (ms)}} \\
        \cline{2-13}
         & \multicolumn{3}{c|}{\textbf{Hemisphere val. set}} & \multicolumn{3}{c?}{\textbf{iHannes test set}} & \multicolumn{3}{c|}{\textbf{Hemisphere val. set}} & \multicolumn{3}{c?}{\textbf{iHannes test set}} \\ 
        \cline{2-13}
        & AP & AP$_{50}$ & AP$_{75}$ & AP & AP$_{50}$ & AP$_{75}$ & AP & AP$_{50}$ & AP$_{75}$ & AP & AP$_{50}$ & AP$_{75}$ \\ \hline

        Mask R-CNN~\cite{he2017} & 62.0 & 87.0 & 69.6 & 19.1 & 38.1 & 17.0 & 50.9 & 78.3 & 55.2 & 11.7 & 28.5 & 7.5 & \textbf{41.0} \\     
        ViTDet~\cite{li2022} & 65.0 & 92.3 & 72.8 & 34.4 & 69.3 & 30.2 & 54.9 & 86.1 & 59.2 & 20.0 & 49.0 & 13.1 & 128.2 \\ \hline      

        \hline
        
        DINOv2Det (Ours) & \textbf{70.1} & \textbf{93.2} & \textbf{79.2} & \textbf{53.1} & \textbf{89.0} & \textbf{57.3} & \textbf{60.5} & \textbf{88.7} & \textbf{67.4} & \textbf{34.9} & \textbf{70.7} & \textbf{29.9} & 54.6 \\  \hline     
        
    \end{tabular}
    \label{table:res_randtext}
    \vspace{-0.3cm}
\end{table*}

\subsection{Object parts segmentation}
\label{sec:exp_object_parts_segmentation}
The proposed DINOv2Det network is compared with Mask R-CNN~\cite{he2017} and ViTDet~\cite{li2022}. For the backbone, we use the smallest DINOv2 model size available, for a fast inference. Instead, Mask R-CNN and ViTDet are based on a ResNet-50 FPN~\cite{lin2017} and a ViT-Base, respectively. The input image size is 640x480px but this is not suitable for DINOv2 since it uses a patch size of 14. Hence, we interpolate the pre-trained weights of the patch embedding filters from 14x14x3 to 16x16x3. We rely on the Detectron2 framework to implement DINOv2Det and to train our models. Mask R-CNN and ViTDet are initialized with the weights provided within the framework. ViTDet is fine-tuned end-to-end while in Mask R-CNN the first two stages of the ResNet-50 are freezed. Regarding DINOv2Det, we first trained from scratch the RPN and RoI heads while keeping DINOv2 (i.e., the backbone) freezed, then fine-tuned the whole network. 
Each model is trained on the synthetic dataset for 22500 iterations 
and we validate it every 1125 iterations on the synthetic validation set presented in Sec.~\ref{sec:synthetic_dataset_generation}. When the training ends, the best performing checkpoint is evaluated on the \textit{iHannes} test set. To assess models performance, we consider the Average Precision (AP) of both the masks and bounding boxes, computed at different Intersection over Union (IoU) of the ground truth with the prediction. Specifically, we use the AP$_{50}$ and AP$_{75}$ with IoU set to 50 and 75, respectively, and the standard COCO AP obtained by averaging the AP over multiple IoUs.
The results and the inference time are shown in Tab.~\ref{table:res_origtext}. Interestingly, Mask R-CNN and DINOv2Det obtain comparable performance on the synthetic validation set. However, DINOv2Det achieves higher results on the \textit{iHannes} test set (+18.0 on the bounding box AP and +13.8 on the Mask AP), proving the effectiveness of adopting DINOv2 features for instance segmentation in zero-shot sim-to-real transfer. The cross-dataset generalization capability of self-supervised features is further confirmed by comparing Mask R-CNN with ViTDet. Indeed, while the former can better fit the training data, the latter retains more general features, resulting in higher performance when moving from the synthetic to the real domain. Finally, we show the AP$_{50}$ and AP$_{75}$ to highlight that the models are generally able to coarsely identify the object parts in the image (e.g., 73.8 for the Mask AP$_{50}$ of DINOv2Det). The main challenge is to precisely draw the mask of the object part, which for some objects is ambiguous since no clear object part boundaries exist, resulting in a performance drop (e.g, from 73.8 for the Mask AP$_{50}$ to 32.9 for the Mask AP$_{75}$ of DINOv2Det).

\noindent\textbf{Texture generalization}. 
We conduct an exploratory study on the generalization to unseen textures for each vision model. For this, we generate a synthetic training set following the procedure in Sec.~\ref{sec:synthetic_dataset_generation}, but applying random textures to objects. Prior to each capture, a texture is uniformly sampled from a set of 15 pre-defined textures and applied to the current object. We trained the models on this dataset and evaluate on the same synthetic validation set and \textit{iHannes} test set as the previous experiment (i.e., where the objects have their original texture). As shown in Tab.~\ref{table:res_randtext}, DINOv2Det exhibits strong generalization to the original (unseen) textures, reporting only a slight drop, i.e., from 36.9 to 34.9 for the Mask AP. This suggests that our model uses the object's shape instead of the texture to classify the parts. We believe this characteristic to be crucial for future works targeting generalization to unseen objects.

%% file: Sections/application.tex
We deploy the proposed wrist control pipeline on the Hannes prosthesis. This arm is equipped with two EMG electrodes placed on the forearm flexor and extensor muscles of the user. These are used to trigger the \textit{rotation} phase and to control the fingers opening-closing during the \textit{grasping} phase. Moreover, the pipeline automatically starts/ends when the user brings the prosthetic arm up/down. An Inertial Measurement Unit (IMU) is used to detect such motion.

The pipeline runs at 15 Hz on a laptop equipped with a NVIDIA RTX 3080. The supplementary video shows a qualitative comparison between the SSC baseline and the proposed vision-based control. Note that during the \textit{transport} phase, we do not run the visual servoing on an object part mask since it would bias the approach. Instead, we use the full object mask. This is obtained by merging the neighboring masks into a single mask, then the closest one to the image center is selected. Subsequently, in the \textit{rotation} phase, the closest object part mask to the image center is selected and its label is used to rotate the wrist accordingly.

%% file: Sections/conclusions.tex
In this work, we addressed the challenge of controlling the wrist of a prosthetic arm during a reach-to-grasp task. Leveraging the shared-autonomy principle, we introduced a novel computer vision-based framework focused on a continuous wrist control followed by predicting the final wrist configuration for grasping. The system has the potential to reduce both the compensatory body movements and the cognitive burden on the user. Such characteristics can be quantified using a motion capture system and fatigue measures such as the pupil dilation. We leave this for future work by testing our system with amputees.

%% file: root.bbl
\begin{thebibliography}{10}
\providecommand{\url}[1]{#1}
\csname url@rmstyle\endcsname
\providecommand{\newblock}{\relax}
\providecommand{\bibinfo}[2]{#2}
\providecommand\BIBentrySTDinterwordspacing{\spaceskip=0pt\relax}
\providecommand\BIBentryALTinterwordstretchfactor{4}
\providecommand\BIBentryALTinterwordspacing{\spaceskip=\fontdimen2\font plus
\BIBentryALTinterwordstretchfactor\fontdimen3\font minus \fontdimen4\font\relax}
\providecommand\BIBforeignlanguage[2]{{%
\expandafter\ifx\csname l@#1\endcsname\relax
\typeout{** WARNING: IEEEtran.bst: No hyphenation pattern has been}%
\typeout{** loaded for the language `#1'. Using the pattern for}%
\typeout{** the default language instead.}%
\else
\language=\csname l@#1\endcsname
\fi
#2}}

\bibitem{chen2023}
Z.~Chen, H.~Min, D.~Wang, Z.~Xia, F.~Sun, and B.~Fang, ``A review of myoelectric control for prosthetic hand manipulation,'' \emph{Biomimetics}, vol.~8, no.~3, 2023.

\bibitem{amsuess2014}
S.~Amsuess, P.~Goebel, B.~Graimann, and D.~Farina, ``Extending mode switching to multiple degrees of freedom in hand prosthesis control is not efficient,'' in \emph{2014 36th Annual International Conference of the IEEE Engineering in Medicine and Biology Society}, 2014, pp. 658--661.

\bibitem{newbury2023deep}
R.~Newbury, M.~Gu, L.~Chumbley, A.~Mousavian, C.~Eppner, J.~Leitner, J.~Bohg, A.~Morales, T.~Asfour, D.~Kragic, \emph{et~al.}, ``Deep learning approaches to grasp synthesis: A review,'' \emph{IEEE Transactions on Robotics}, vol.~39, no.~5, pp. 3994--4015, 2023.

\bibitem{gardner2020}
M.~Gardner, C.~S. Mancero~Castillo, S.~Wilson, D.~Farina, E.~Burdet, B.~C. Khoo, S.~F. Atashzar, and R.~Vaidyanathan, ``A multimodal intention detection sensor suite for shared autonomy of upper-limb robotic prostheses,'' \emph{Sensors}, vol.~20, no.~21, p. 6097, 2020.

\bibitem{vasile2022}
F.~Vasile, E.~Maiettini, G.~Pasquale, A.~Florio, N.~Boccardo, and L.~Natale, ``Grasp pre-shape selection by synthetic training: Eye-in-hand shared control on the hannes prosthesis,'' in \emph{2022 IEEE/RSJ International Conference on Intelligent Robots and Systems (IROS)}.\hskip 1em plus 0.5em minus 0.4em\relax IEEE, 2022, pp. 13\,112--13\,119.

\bibitem{starke2022}
J.~Starke, P.~Weiner, M.~Crell, and T.~Asfour, ``Semi-autonomous control of prosthetic hands based on multimodal sensing, human grasp demonstration and user intention,'' \emph{Robotics and Autonomous Systems}, vol. 154, p. 104123, 2022.

\bibitem{oquab2023}
M.~Oquab, T.~Darcet, T.~Moutakanni, H.~Vo, M.~Szafraniec, V.~Khalidov, P.~Fernandez, D.~Haziza, F.~Massa, A.~El-Nouby, \emph{et~al.}, ``Dinov2: Learning robust visual features without supervision,'' \emph{arXiv preprint arXiv:2304.07193}, 2023.

\bibitem{he2017}
K.~He, G.~Gkioxari, P.~Doll{\'a}r, and R.~Girshick, ``Mask r-cnn,'' in \emph{Proceedings of the IEEE international conference on computer vision}, 2017, pp. 2961--2969.

\bibitem{laffranchi2020hannes}
M.~Laffranchi, N.~Boccardo, S.~Traverso, L.~Lombardi, M.~Canepa, A.~Lince, M.~Semprini, J.~A. Saglia, A.~Naceri, R.~Sacchetti, E.~Gruppioni, and L.~D. Michieli, ``The hannes hand prosthesis replicates the key biological properties of the human hand,'' \emph{Science Robotics}, vol.~5, no.~46, p. eabb0467, 2020.

\bibitem{castro2022semi}
M.~N. Castro and S.~Dosen, ``Semi-autonomous prosthesis control using minimal depth information and vibrotactile feedback,'' \emph{arXiv:2210.00541}, 2022.

\bibitem{shi2023hand}
X.~Shi, W.~Guo, W.~Xu, and X.~Sheng, ``Hand grasp pose prediction based on motion prior field,'' \emph{Biomimetics}, vol.~8, no.~2, p. 250, 2023.

\bibitem{robotics12060152}
\BIBentryALTinterwordspacing
G.~Cirelli, C.~Tamantini, L.~P. Cordella, and F.~Cordella, ``A semiautonomous control strategy based on computer vision for a hand–wrist prosthesis,'' \emph{Robotics}, vol.~12, no.~6, 2023. [Online]. Available: \url{https://www.mdpi.com/2218-6581/12/6/152}
\BIBentrySTDinterwordspacing

\bibitem{shi2022target}
X.~Shi, W.~Xu, W.~Guo, and X.~Sheng, ``Target prediction and temporal localization of grasping action for vision-assisted prosthetic hand,'' in \emph{2022 IEEE International Conference on Robotics and Biomimetics (ROBIO)}.\hskip 1em plus 0.5em minus 0.4em\relax IEEE, 2022, pp. 285--290.

\bibitem{chaumette2006}
F.~Chaumette and S.~Hutchinson, ``Visual servo control. i. basic approaches,'' \emph{IEEE Robotics \& Automation Magazine}, vol.~13, no.~4, pp. 82--90, 2006.

\bibitem{nguyen2017}
A.~Nguyen, D.~Kanoulas, D.~G. Caldwell, and N.~G. Tsagarakis, ``Object-based affordances detection with convolutional neural networks and dense conditional random fields,'' in \emph{2017 IEEE/RSJ International Conference on Intelligent Robots and Systems (IROS)}, 2017, pp. 5908--5915.

\bibitem{myers2015}
A.~Myers, C.~L. Teo, C.~Fermüller, and Y.~Aloimonos, ``Affordance detection of tool parts from geometric features,'' in \emph{2015 IEEE International Conference on Robotics and Automation (ICRA)}, 2015, pp. 1374--1381.

\bibitem{do2018affordancenet}
T.-T. Do, A.~Nguyen, and I.~Reid, ``Affordancenet: An end-to-end deep learning approach for object affordance detection,'' in \emph{2018 IEEE international conference on robotics and automation (ICRA)}.\hskip 1em plus 0.5em minus 0.4em\relax IEEE, 2018, pp. 5882--5889.

\bibitem{gentilucci1991}
M.~Gentilucci, U.~Castiello, M.~Corradini, M.~Scarpa, C.~Umilta, and G.~Rizzolatti, ``Influence of different types of grasping on the transport component of prehension movements,'' \emph{Neuropsychologia}, vol.~29, no.~5, pp. 361--378, 1991.

\bibitem{chaumette2007}
F.~Chaumette and S.~Hutchinson, ``Visual servo control. ii. advanced approaches [tutorial],'' \emph{IEEE Robotics \& Automation Magazine}, vol.~14, no.~1, pp. 109--118, 2007.

\bibitem{corke2023}
P.~Corke, \emph{Robotics, Vision and Control: Fundamental Algorithms in Python}.\hskip 1em plus 0.5em minus 0.4em\relax Springer Nature, 2023, vol. 146.

\bibitem{dosovitskiy2020}
A.~Dosovitskiy, L.~Beyer, A.~Kolesnikov, D.~Weissenborn, X.~Zhai, T.~Unterthiner, M.~Dehghani, M.~Minderer, G.~Heigold, S.~Gelly, \emph{et~al.}, ``An image is worth 16x16 words: Transformers for image recognition at scale,'' in \emph{International Conference on Learning Representations}, 2021.

\bibitem{li2022}
Y.~Li, H.~Mao, R.~Girshick, and K.~He, ``Exploring plain vision transformer backbones for object detection,'' in \emph{European conference on computer vision}.\hskip 1em plus 0.5em minus 0.4em\relax Springer, 2022, pp. 280--296.

\bibitem{calli2015}
B.~Calli, A.~Walsman, A.~Singh, S.~Srinivasa, P.~Abbeel, and A.~M. Dollar, ``Benchmarking in manipulation research: The ycb object and model set and benchmarking protocols,'' \emph{arXiv:1502.03143}, 2015.

\bibitem{wang2021}
G.~Wang, F.~Manhardt, F.~Tombari, and X.~Ji, ``Gdr-net: Geometry-guided direct regression network for monocular 6d object pose estimation,'' in \emph{Proceedings of the IEEE/CVF Conference on Computer Vision and Pattern Recognition}, 2021, pp. 16\,611--16\,621.

\bibitem{park2017colored}
J.~Park, Q.-Y. Zhou, and V.~Koltun, ``Colored point cloud registration revisited,'' in \emph{2017 IEEE International Conference on Computer Vision (ICCV)}, 2017, pp. 143--152.

\bibitem{boccardowrist}
N.~Boccardo, M.~Canepa, S.~Stedman, L.~Lombardi, A.~Marinelli, D.~D. Domenico, R.~Galviati, E.~Gruppioni, L.~D. Michieli, and M.~Laffranchi, ``Development of a 2-dofs actuated wrist for enhancing the dexterity of myoelectric hands,'' \emph{IEEE Transactions on Medical Robotics and Bionics}, vol.~6, no.~1, pp. 257--270, 2024.

\bibitem{di2021hannes}
D.~Di~Domenico, A.~Marinelli, N.~Boccardo, M.~Semprini, L.~Lombardi, M.~Canepa, S.~Stedman, A.~D. Bellingegni, M.~Chiappalone, E.~Gruppioni, M.~Laffranchi, and L.~De~Michieli, ``Hannes prosthesis control based on regression machine learning algorithms,'' in \emph{2021 IEEE/RSJ International Conference on Intelligent Robots and Systems (IROS)}, 2021, pp. 5997--6002.

\bibitem{lin2017}
T.-Y. Lin, P.~Doll{\'a}r, R.~Girshick, K.~He, B.~Hariharan, and S.~Belongie, ``Feature pyramid networks for object detection,'' in \emph{Proceedings of the IEEE conference on computer vision and pattern recognition}, 2017, pp. 2117--2125.

\end{thebibliography}
